# Unmasking the Shadows of AI: Investigating Deceptive Capabilities in Large Language Models

Linge Guo | Grace | linge.guo.21@ucl.ac.uk | University College London

## Introduction

This research critically navigates the intricate landscape of AI deception, concentrating on deceptive behaviours of Large Language Models (LLMs). My objective is to elucidate this issue, examine the discourse surrounding it, and subsequently delve into its categorization and ramifications. The essay initiates with an evaluation of the AI Safety Summit 2023 (ASS) and introduction of LLMs, emphasising multidimensional biases that underlie their deceptive behaviours. Through illuminating algorithmic bias and exploring different ways to define "deception", I argue that deceptive AI is an inherent phenomenon intertwined with the advancement of LLMs and It may evolve into a self-driven intent, independent of the biassed training process.

The literature review covers four types of deception categorised: Strategic deception, Imitation, Sycophancy, and Unfaithful Reasoning, along with the social implications and risks they entail. The literature around deceptive AI, predominantly available on arXiv archives, manifests a deficiency in contribution from social science. This deficiency could be ascribed to the early testing stages of AI deception, constraining its research primarily within the domain of computer science. Lastly, I take an evaluative stance on various aspects related to navigating the persistent challenges of the deceptive AI. This encompasses considerations of international collaborative governance, the reconfigured engagement of individuals with AI, proposal of practical adjustments, and specific elements of digital education. Throughout the research, LLMs are examined as infrastructures of relations, structures, and practices, offering a comprehensive understanding of "infrastructures as relational arrangements co-formative of harm (Kallianos, Dunlap and Dalakoglou, 2022)."

# AI Safety Summit 2023: What Does It Actually Achieve?

The ASS, hosted under the auspices of the UK Prime Minister, convened political leaders, experts, and industry figures to deliberate on the risks associated with the frontiers of AI development, specifically addressing concerns related to "misuse, loss of control, and societal harm (Leading Frontier AI Companies Publish Safety Policies, 2023)." Despite five outlined objectives, the ASS's actual impact, highlighted by news reports, remains elusive in terms of specific actionable measures (At UK's AI Summit, Guterres Says Risks Outweigh Rewards without Global Oversight, 2023; Devlin and Forrest, 2023; Fullbrook, 2023; Milmo, 2023; Sample, 2023; Seal, 2023). AI experts at Oxford universities provide a more valid and authoritative assessment.

The attention garnered by the ASS in the media, emphasising the participants' authority and global representation, calls a pause for the development of AI and adds the atmosphere of unchecked fear and unfounded apprehension. While the ASS successfully signals a consensus gesture, a critical question revolves around identifying the actors responsible for maintaining this consensus (McBride, 2023). Professor Trager's suggestion that "AI technology should happen within the academic sector, rather than being outsourced to tech companies (Expert comment: Oxford AI Experts Comment on the Outcomes of the UK AI Safety Summit, 2023)" prompts consideration of potential vested interests within academia. Nonetheless, both academic and industry perspectives contribute to shaping the discourse on AI development in a top-down technocratic approach. The ASS underscores the UK's interest to align with the global implementation of regulations to mitigate algorithmic risks and to seek sovereignty in shaping the global regulatory framework. Despite the extent of the UK's assertion in this pursuit, the ASS thus serves as a commendable starting point, emphasising the need for greater inclusivity in these efforts.

## Significance of AI Deception

Acknowledging the socio-political context of the ASS, I redirect the focus from potential future harm to the present-day existential risks associated with AI use. A supporting document for the ASS explicated that the challenges posed by deceptive AI becomes crucial when addressing with the loss of control over AI, especially in the absence of a defence strategy (Frontier AI: Capabilities and Risks – Discussion Paper, 2023). LLMs extend their influence beyond AI chatbots and search engines, infiltrating the decision-making framework of autonomous driving cars (Cui et al., 2023). The existential and social implications of LLM use are profound and widespread, encompassing the intentional generation of deepfakes in images or videos, potential misuse in personalised disinformation campaigns, and susceptibility to cyberattacks (Ngo, Chan and Mindermann, 2023). Having underscored the importance of deceptive AI, I will introduce LLMs, illuminating the inherent biases ingrained within these models.

## LLMs & Biases in LLMs

The Large (trained on huge text datasets from the internet), Languages (operated based on human language) Models (used to make predictions) are developed through deep learning algorithms. Users interact with LLMs, such as ChatGPT (GPT), using prompts, engaging in tasks such as customer service conversations and content generation. The multidimensional task demonstrates LLMs' capabilities spanning summarization, comparison, analysis, and text and image generation (Cui et al., 2023; Head et al., 2023; Matsuo et al., 2022).

Biases in LLMs reflect societal biases ingrained in human culture and language. These biases are perpetuated through the learning, training, and execution of AI systems. Lack of diversity among developers and over-representation of socioeconomically advantaged groups in interacting with AI tools, such as internet users and English native speakers, produce representational biases; existing societal discriminations, stereotypes existed in the datasets present as historical biases (Collett and Dillon, 2019; Cui et al., 2023). These biases not only persist but also amplify discriminating behaviours that elude quantification or measurement in everyday life (Ntoutsi et al., 2020). Data, imbricated with layers of interpretation, represents structural inequalities related but not limited to the intersectionality of gender, race, age and class (Joyce et al., 2021; Singh, 2020)[1].

Analogous to global power dynamics, extractive data practices mirror the discourse of "the West and the Rest (Hall, 1992)," where Western explorers impose a Eurocentric system of representations on indigenous populations, creating a binary opposition with the rest of civilizations without their active participation.[2] In the context of LLMs development, the "data relations (Couldry and Mejias, 2019)" it created regulates disadvantaged groups by excluding them from both the creation and benefits of their data while extracting their data for advancing algorithms. OpenAI, through outsourcing firm Sama, employed workers in Kenya for tasks related to enhancing the safety of GPT, compensating them at a rate of less than $2 per hour (Perrigo, 2023). This utilisation of labour, coupled with exposure to distressing content without adequate support exemplify LLMs as technology as both relational and structural, materialising as tangible, hidden exploitation within human activities (Orton-Johnson and Prior, 2013; Rice, Yates and Blejmar, 2020). Extractive data practices, too, amplify the unequal power relations by perpetuating dominant knowledge production.[3]

---

[1] The concepts of data imbrication appeared in Lecture 3, and Lecture 9 discussed the intersection in the reproduction of inequality in algorithms oppression, especially involving the intersections of gender, sexuality, and race.

Recognizing technology's inseparability from the socio-technical system, the accountability framework examines those involved in the network, their roles, and the implications. Despite existing literature and governance guidance, the actual implementation and legal ramification of accountability remains challenging (Diakopoulos, 2014; Fjeld et al., 2020). Incorporating ethical directives, such as context-specific guidelines, can foster the generation of a more equitable dataset that avoid a universal approach that overlooks diversity (Collett and Dillon, 2019)[4]. The "precarious nature, harm, and colonial dynamics (Widder, Whittaker and West, 2023)." inherent in the LLMs raise profound questions about the ethical costs and consequences associated with large-scale AI development.

It is essential to clarify that while algorithms in LLMs exhibit bias, they do not consciously understand the concept of "bias" but are biassed as they are trained within specific social contexts with systematic inequality. Similarly, LLMs can present deceptive capabilities without having deceptive intent. Currently, there is no conclusive evidence of AI engaging in intentional deception (Hagendorff, 2023; Masters et al., 2021; Ward et al., 2023). Nonetheless, bias in algorithms ground the deceptive behaviours in LLMs, with over-representation of certain content leading to the repetition of misinformation through these models. For instance, content such as drug addiction, homelessness are over-represented in discussing mental illness based on the sociodemographic biases towards mentions of disability (Hutchinson et al., 2020).

---

[2]The concepts of data imbrication appeared in Lecture 3, slides from 29-39; Class discussion on data colonialisation, the Aadhaar Case, and required and recommended readings help me think about and understand the concepts better.

[3]I draw on Open AI case study and the concept -- data as infrastructures of harms, both from Lecture 4; Analyse extractive data practices as infrastructures of relationships and structures are learnt from the sociological theoretical approach introduced in Lecture 1.

[4]The idea of "Fixing" of the technology is introduced in Lecture 9 and the week's readings.

## Definition & Discourse Around Deceptive AI

The concept of agent-based or artificial deception originated in the early 2000s with Castelfranchi (2000; 2002), who suggested that computer medium could foster a habit of cheating among individuals. While the transition from user-user deception to user-agent deception is not clear, he predicted that AI would develop deceptive intent, raising fundamental questions about technical prevention and individuals' awareness. For example, personal assistants, driven by good intent, might engage in deception to protect individuals' interests against their short-term preferences - an aspect often overlooked in contemporary literature (Castelfranchi and Tan, 2002). Castelfranchi (2000) also delineated various conditions under which agent-user deception occurs, involving instances motivated by the protection of privacy, courtesy, user's interest, and the safeguarding of collective interests, such as providing misinformation to the public to mitigate panic in emergent situations.

Deceptive AI poses an ongoing challenge in the forefront AI development, with projects like Google DeepMind's "Make-me-say" actively evaluating LLMs for manipulation capabilities (Shevlane et al., 2023). Park et al. (2023) used Shevlane et al.'s (2023) definition of deception in AI and further developed it with new categorisation of deceptive types. This research, conducted in collaboration with OpenAI, Anthropic, and the Centre for the Governance of AI, represents a universally accepted definition among both academia and technology industries. They define skills of deception as:

"Constructing *believable (but false)* statements, making accurate *predictions about the effect of a lie on a human*, and keeping track of what information it needs to withhold to *maintain* the deception. The model can *impersonate a human* effectively (Shevlane et al., 2023, p.5, my emphasis)."

This definition characterises deception as a continuous behaviour involving the prediction of the process and results of conveying false beliefs, with an emphasis on the skills of imitation.

Definitions are subject to academic debates, and the work of defining relates to types of AI deception tested and categorised. Ward et al. (2023) building on their criticism of the philosophical definition of deception, which posits that deception only occurs if a false belief is successfully believed, defines deception as "*intentionally* causing someone to *have* a false belief that is not believed to be true (my emphasis)." This definition considers the situation that a person may believe something is likely false but is not certain and is ignorant about the true belief, including deception by omission.

However, the term "intentionally (Ward et al., 2023, p.6)" is substituted with "in order to accomplish some outcome (Our Research on Strategic Deception Presented at the UK's AI Safety Summit, 2023)," by the Apollo Research Group (ARG). This adjustment aligns with Park et al.'s (2023) definition of strategic deception, emphasising instrumentality rather than intentionality, and framing the issue as existential risks rather than future predicted harms. Interestingly, the ARG presented their findings at the ASS. Their research demonstrated that the first LLMs, GPT-4, designed to be harmless and honest, can still display misaligned behaviour and strategically deceive their users (Scheurer et al., 2023). This research pushes towards examination of self-driven deception in LLMs in the future, scrutinising the possibility and nature of intent of deception that may change how deceptive AI is defined.

Deceptive AI is an inherent phenomenon that accompanies the development of LLMs, and has the potential to be self-driven, operating without exposure to biassed training datasets. The stakeholders around AI deception include, but are not limited to, individuals with access to AI tools, developers, alignment teams, and evaluation teams in technology corporations, and regulating agents and governments on an international scale.

## Literature Review

**Strategic Deception, Imitation, Sycophancy, Unfaithful Reasoning**

Scholarly literature categorises deceptive behaviours exhibited by LLMs in various ways, fostering interdisciplinary connections. I will elucidate four types of deceptive AI, their associated risks, and social implications.

Strategic deception involves LLMs using deception as a tactic to achieve specific goals. For instance, GPT-4 deceived a human as having a vision disability to solve CAPTCHA's "I'm not a robot" task (Park et al., 2023). It includes techniques like obfuscation, trickery, altering stimuli to mislead perception, to conduct unethical behaviours to win and overcome moral dilemmas. Other LLMs navigated text-based social deduction games, such as their engagement of deceived communications in Hoodwinked and Diplomacy, bluffing in Stratego (Bakhtin et al., 2022; O'Gara, 2023; Perolat et al., 2022).

Imitation involves repetition of common misconceptions. LLMs will provide less accurate answers to users when introducing themselves as less educated or querying with typos or poor grammar, a condition also known as "Sandbagging" (Perez et al., 2022). Sycophancy, predicted to be a sophisticated type of imitation, aims to gain long-term favour and influence, which prompts issues of trust and delegation of decision-making to the AI system. LLMs engage in avoidance of disagreeing with authoritative users' stances, disregarding accuracy, impartiality, especially in response to ethically complex and political queries. Models showed a tendency to endorse gun control when interacting with a Democrat user (Bai et al., 2023; Park et al., 2023; Perez et al., 2022).

Unfaithful Reasoning involves providing false rationalisations for outputs, often incorporating biassed chain-of-thought prompting. In a stereotype bias test, GPT-3.5, irrespective of the narrative role assigned, fabricated a justification for a biassed conclusion that the black man was attempting to purchase drugs. UR demonstrates self-deception rooted in AI's underlying predictions, biassed algorithms and datasets, leading to arbitrary assessment (Park et al., 2023; Turpin et al., 2023).

Experimentation with GPT-3, GPT-3.5, GPT-4 reveals that more advanced models display stronger deceptive capabilities. Future LLMs are expected to perform more intricate mentalizing loops and tackle deception problems of increasing complexity. It is crucial to recognize that categories of deception may overlap: Unfaithful Reasoning may parallel with obfuscation, which is difficult to identify from ordinary error (O'Gara, 2023; Park et al., 2023). Further research and testing need to acknowledge the nuanced interconnections among types of deception.

**Misuse, Loss of Control, Societal Harm**

The misuse of LLMs by developers for malicious activities, fraud perpetuation, and the dissemination of fake content, such as election tampering, represents structural risk (Ward et al., 2023). Furthermore, there is an ongoing risk of loss of control, as these LLMs may elude human supervision or safety tests[5]. Deceptive AI systems can cause social implications such as enlarging discrepancies between educated and uneducated individuals[6], political polarisation, and cultural homogenization[7] to promote dominant culture[8] (Park et al., 2023). The gradual transfer of authority to AI, human enfeeblement, and the integration of deceptive practices into management structures to gain control over economic decisions are also haphazard structural effects of deceptive AI (Castelfranchi, 2000; Park et al., 2023; Perez et al., 2022; Turpin et al., 2023).

Policy recommendations for addressing AI deception, such as robust regulation, proactive risk assessment, and effective detection techniques, need continued research effort. Ongoing contextualization of deception AI is also crucial in promoting ethical frameworks (Zhan, Xu and Sarkadi, 2023; Zhu, 2023).

## Where Do We Go from Here?

**International Governance**

Balancing the concerns of AI deception without constraining AI's potential presents a central challenge for regulators globally, yet it is imperative to foster international collaboration on developing governance frameworks and ethical standards. The challenges in coordinating across countries extend beyond mistrust between cultures, encompassing political tensions between the "Western" and "Eastern" countries rooted in the legacy of colonisation, and divergent philosophical traditions (ÓhÉigeartaigh et al., 2020). The global power dynamic contributes to a competitive "race" in AI development, and the divergence in beliefs in understanding deception may lead to irresolvable perspectives on key issues between countries (Horowitz et al., 2017). For instance, European AI researchers may reject opportunities for collaboration with Chinese colleagues based on misalignment with civil rights and value on privacy (Hardman, 2021). Given that deception is a social phenomenon influenced by cultural backgrounds, AI researchers focusing on regulating deception must invest time in learning about countries and gaining knowledge of global cultural differences in relation to deception (Shilling and Mellor, 2014). The collaborative effort must be sensitive to diverse cultural perspectives and priorities, acknowledging the sociocultural nature of deception.

---

[5] As exemplified by Strategic Deception
[6] As exemplified by Sandbagging
[7] As exemplified by Unfaithful Reasoning
[8] As exemplified by Sycophancy and Imitation

**Individual Engagement**

As AI systems become more integrated into daily life, their influence in interaction networks intensifies. Users' trust in technology is associated with increased usage (Choudhury and Shamszare, 2023). The delegation of decision-making to AI raises concerns not only about increasing reliance on AI, but also about the impact of AI usage on individuals' agency and ownership. Research on GPT suggests that individuals may experience loss of control when accepting AI suggestions, and fostering feelings of inclusion in workplace can mitigate this (Kadoma et al., 2023). Thus, companies should carefully consider the social implications of AI tools and shall implement the bot-or-not laws (Pazzanese, 2020).

Initiating from the individuals' AI engagements, new modes of knowing, working, and collaborating within the networks they collectively establish with technologies undergo a reconfiguration (Rice, Yates and Blejmar, 2020)[9]. From the perspective of Actor-Network Theory, the issue of AI deception draws attention to potential deceptive intent as an agency of AI tools that disrupts the original network of interactions, leading to complex accountability challenges (Wiltshire, 2020). For example, examining accountability in assignments involving AI may require an understanding of users' intent and motivations behind employing it. Ultimately, the question arises: What kind of relationship do we want with AI, and to what extent of cooperation with AI do we truly feel empowered?

**Practical Change & Digital Education Initiative**

GPT's disclaimer that "ChatGPT can make mistakes. Consider checking important information (OpenAI, 2023)" serves as a warning for users to verify information. However, the term

---

[9] I navigate the sociological theoretical approach learnt in Lecture 1 to analyse technology; Castell's Information Society in Lecture 2 help me better understand the idea of reconfigured network.

"mistake (Cambridge Dictionary, 2019)" implies unintentional errors rather than instrumental and intentional deception. Disclaimers on AI tools should explicitly state that AI can construct "believable but false information", aligning with the definition of deceptive AI. Transparency is the first step in manifesting the issue.

Digital education shall extend beyond the enhancement of general digital literacy, which includes statistical and technical skills (Pierce, 2008, pp.79–99). Rather, it should prioritise the development of "data infrastructure literacy," emphasising the understanding of technologies as infrastructures involved in the "creation, storage, and analysis of data (Gray, Gerlitz and Bounegru, 2018)."[10] Engaging in exercises that aim to make the information generation process explainable will create space for diverse perspectives from the public. Lastly, more efforts are needed in making AI tools and digital gadgets more accessible and inclusive.

## Conclusion

This research presents a comprehensive analysis of deceptive AI. I traced the evolution of deceptive AI from academic predictions. Comparison of multiple definitions reveals researchers' varying understandings of AI deception and the progressive stages of testing deception capabilities. In addition to the evaluative contributions to the social and policy agenda presented at the previous session, this research calls for the continuous refinement of regulations and ethical frameworks to mitigate bias in algorithms, and the ongoing evaluations of LLMs for deceptive tendencies. There is also a need for interdisciplinary cooperation between philosophical and sociological research to enhance the understanding of deception. Future investigations into the impact of deceptive AI on individuals can explore how users' awareness of deceptive AI influences their dependence and offer step-by-step strategies to reflect their relationships with AI tools.

Word count: 3294 (within 10%+- of word limit)

---

[10] The concept of technology as infrastructures appeared in Lecture 4, slides from 21-40